\documentclass[12pt, leqno]{article}
\usepackage{amsmath}
\usepackage{amsfonts}
\usepackage{sabzikar}
\usepackage{oxford3}
\usepackage[notref, notcite]{showkeys}
\usepackage{color}
\usepackage{float}
\usepackage{graphicx}
\usepackage{comment}

\makeatletter
\g@addto@macro{\normalsize}{
    \setlength{\abovedisplayskip}{2pt}
    \setlength{\abovedisplayshortskip}{0pt}
    \setlength{\belowdisplayskip}{4pt}
    \setlength{\belowdisplayshortskip}{4pt}}
\makeatother

\newcommand \cB{\mathcal B}

 \newcommand \mbR{\mathbb  R}

\begin{document}

\begin{center}
\Large \textbf{Continuous Delayed-Memory Stochastic Gradient Descent and Continuous-Time Reinforcement Learning from History of Astrophysical Time Series Studies}

\bigskip

Juncheng Yi\\[2pt]
{ccyi3@iastate.edu}\\[2pt]
{\small May 8th 2026}
\end{center}

\bigskip

\begin{abstract}
\noindent
Quasars are luminous objects in the universe that exhibit stochastic brightness variations encoding information about the supermassive black holes powering them, and modeling these variations from ground-based survey data time series, known as light curves, is a statistical challenge\cite{ivezic_macleod_quasars}. This paper reviews how stochastic differential equations (SDEs) have been adapted with neural network parameterizations to overcome this challenge in history: \cite{ivezic_macleod_quasars} \cite{Fagin_2024}, \cite{10.1145/3746252.3760805}. Following \cite{10.1145/3746252.3760805}, we create the Continuous-Delayed-Memory Stochastic Gradient Descent which depend on the past state of the discrete iteration process. We performed the simulation on some 2-dimensional landscape and observed some wider-exploration and more precise convergent behavior compared to Vanilla SGD by adjusting hyperparameters. Besides, we proposed a reinforcement learning structure with continuous time policy gradients for exploratory policies without solving HJB PDE, and we show that its optimality conditions recover the Gibbs policy of \cite{JMLR:v21:19-144}.

\medskip\noindent {\it Keywords:}
Stochastic process, Stochastic gradient descent, Continuous-Delayed-Memory Stochastic Gradient Descent, Stochastic Delay Differential Equation, Reinforcement Learning, Adjoint method

\end{abstract}

\bigskip

\section{Introduction of Pre-Neural SDE Methods for Astrophysical Quasar Analysis }
\label{s:intro}

Three mathematical objects sit at the heart of this project:
\begin{enumerate}
\item the \emph{Ornstein--Uhlenbeck SDE}, which has modeled quasar optical variability since \citetext{Kelly_2009}
\item \emph{stochastic gradient descent}, the workhorse optimization algorithm of modern
machine learning \cite{doi:10.1137/16M1080173}
\item the \emph{stochastic adjoint method} (a backward SDE) that makes training Neural SDEs
practical \cite{pmlr-v108-li20i}
\end{enumerate}
At first glance, these live in different worlds. The OU process is a physical model; SGD is a numerical algorithm; the adjoint is a tool for automatic differentiation. The central observation that structures this project is that \emph{all three are instances of the same idea}: a state (or parameter) vector evolving in time under a deterministic drift plus a noise term, with the noise either injected physically (OU), introduced by mini-batch sampling (SGD), or inherited from the forward Brownian path (the adjoint).

\medskip\noindent Our literature review started with the modeling of quasar light curves from irregularly sampled photometric surveys. Quasars are powered by accretion onto supermassive black holes and exhibit stochastic brightness variations whose statistics encode information about the central engine. The result of \citetext{Kelly_2009} was that quasar optical variability is well described by a Damped Random Walk (DRW), mathematically, the OU SDE
\smallskip
\beq\label{e:OU}
\mathrm{d} X(t) \;=\; -\frac{1}{\tg}\bigl(X(t) - \mu\bigr)\,\mathrm{d} t \;+\; \sg\, \mathrm{d} W(t)
\eeq
\smallskip
where $\tg$ is a damping timescale, $\mu$ is the long-run mean magnitude, $\sg$ is a short-term volatility, and $W(t)$ is a standard Wiener process. The DRW is Gaussian, Markov, and stationary, and it admits the exact transition density
\smallskip
\beq\label{e:OUtransition}
X(t+\Dg t)\,\big|\,X(t)
\;\sim\;
\mathcal{N}\!\lp\mu + \bigl(X(t)-\mu\bigr) e^{-\Dg t/\tg},\;
\frac{\sg^{2}\tg}{2}\bigl(1-e^{-2\Dg t/\tg}\bigr)\rp
\eeq
\smallskip
So the likelihood of an irregularly sampled light curve can be written exactly without any interpolation. This characteristic made ~\refeq{OU} the de-facto standard for time-domain astrophysics for a decade.

\medskip\noindent However, the OU/DRW model has several well-documented limitations. The following list documents some of those limitations, some extensions to other modelling methods and why they fail in this problem.

\medskip\noindent{\bf PSD slope mismatch}\;
The OU process has a Lorentzian power spectral density $P(f)\propto 1/(f_{0}^{2}+f^{2})$, giving a spectral slope of exactly $-2$ at high frequencies. \citetext{Mushotzky_2011} analysed four AGN observed at 30-minute cadence by the \emph{Kepler} space telescope and found PSD slopes ranging from $-2.6$ to $-3.3$, significantly steeper than predicted by DRW. \citetext{10.1093/mnras/stv1230} extended the analysis to 20 \emph{Kepler} AGN and found fewer than half consistent with DRW. Ground-based surveys with their sparser sampling had masked this mismatch.

\medskip\noindent{\bf Linearity and single-band limitation}\;
By Doob's theorem, the OU process is the unique process that is simultaneously Gaussian, Markov, and stationary. Analytic tractability thus comes at the cost of linear dynamics and a single output. Real quasar variability involves nonlinear physical processes (accretion disk instabilities, corona--disk interactions) and multi-band correlations with inter-band time lags that a single-output OU process cannot capture.

\medskip\noindent{\bf CARMA models}\;
\citetext{Kelly_2014} introduced continuous-time autoregressive moving average (CARMA) models, which generalize DRW to higher-order linear SDEs driven by a common Brownian motion. CARMA$(2,1)$ (the damped harmonic oscillator) fits many AGN light curves better than DRW, but the family remains linear and parametric: its PSD is a rational function (a sum of Lorentzians).

\medskip\noindent{\bf Discrete-time deep learning models}\;
Recurrent networks (RNNs, LSTMs, GRUs) operate on discrete, regularly spaced time steps. Handling irregular sampling requires interpolation, binning, or imputation, all of which can introduce artifacts. More fundamentally, these models are deterministic mappings from input sequences to outputs; they are not generative models for the underlying process and do not quantify uncertainty in the latent dynamics.

\medskip\noindent{\bf Neural ODEs: continuous but deterministic}\;
Neural ODEs \cite{10.5555/3327757.3327764} model the latent state as the solution of $\mathrm{d} z/\mathrm{d} t = f_{\thg}(z(t),t)$ for a neural network $f_{\thg}$, integrated by a numerical solver that can be queried at arbitrary times. Neural ODEs are deterministic: given an initial condition, the trajectory is fully determined. It makes them unsuitable for systems where different realizations from the same initial condition produce different trajectories---precisely the situation for quasar variability, where the stochasticity is physical (turbulence in the accretion flow) rather than merely observational.

\medskip\noindent These models' limitations collectively motivate Neural SDEs, which combine the continuous-time, irregular-sampling-compatible framework of Neural ODEs with the stochastic dynamics that are physically appropriate for quasar variability, while using neural networks to move beyond linear parametric constraints.

\medskip\noindent With the Rubin Observatory's Legacy Survey of Space and Time (LSST), which delivers light curves for roughly $10^{7}$ quasars in six photometric bands, the DRW is no longer adequate: it is linear, single-band, and has a power spectral slope that is systematically wrong at short timescales \cite{Mushotzky_2011,10.1093/mnras/stv1230}. Neural SDEs replace the fixed parametric drift and diffusion of~\refeq{OU} with neural networks while retaining the continuous-time stochastic framework.

\medskip\noindent{\bf Organization}\;
Sections~\ref{s:fagin}  reviews the astrophysical applications of latent neural SDE from \citetext{Fagin_2024}, and Sections~\ref{s:sdde} reviews \citetext{10.1145/3746252.3760805} and illustrate how it progresses from \citetext{Fagin_2024} by considering past state in stochastic process. Section~\ref{s:sdgd} derives Continuous-Delayed-Memory SGD and performs its simulations in 2D parameter space. Section~\ref{s:rl} develops the continuous-time RL story, culminating in the Exploratory Backward Stratonovich SDE. To conclude, Section~\ref{s:future} enumerates several compelling directions for future research.

\section{Latent SDEs for Quasar Light Curves: Fagin et al.\ (2024)}
\label{s:fagin}

\citetext{Fagin_2024} applied the latent SDE framework to astrophysical time series for the first time. Their model simultaneously (i)~reconstructs multi-band quasar light curves across seasonal gaps and (ii)~infers physical properties of the accreting black hole.

\subsection{Problem Formulation}

For a single quasar, the data consist of irregularly sampled magnitudes in $B=6$ bands ($u,g,r,i,z,y$) over a 10-year LSST baseline: $\{(t_{i}, x_{t_{i}}^{(b)}, \sg_{t_{i}}^{(b)}, m_{t_{i}}^{(b)})\}$, where $x$ is the measured magnitude, $\sg$ is the photometric error, and $m^{(b)}\in\{0,1\}$ is a band-specific observation mask. Spacings between successive $t_{i}$ are irregular, with seasonal gaps of $\sim 6$ months.

\medskip\noindent The physical parameters to infer are the black hole mass $\log_{10}(M_{\rm BH}/M_{\odot})\in[7,10]$, the disk inclination $i$, the temperature profile exponent $\bg$ (where $T(r)\propto r^{-\bg}$; standard thin-disk theory predicts $\bg=3/4$), and the DRW parameters $\tg$ and $\mathrm{SF}_{\infty}$.

\subsection{Model Architecture}

\begin{figure}[h!]
    \centering
    \includegraphics[width=0.71\linewidth]{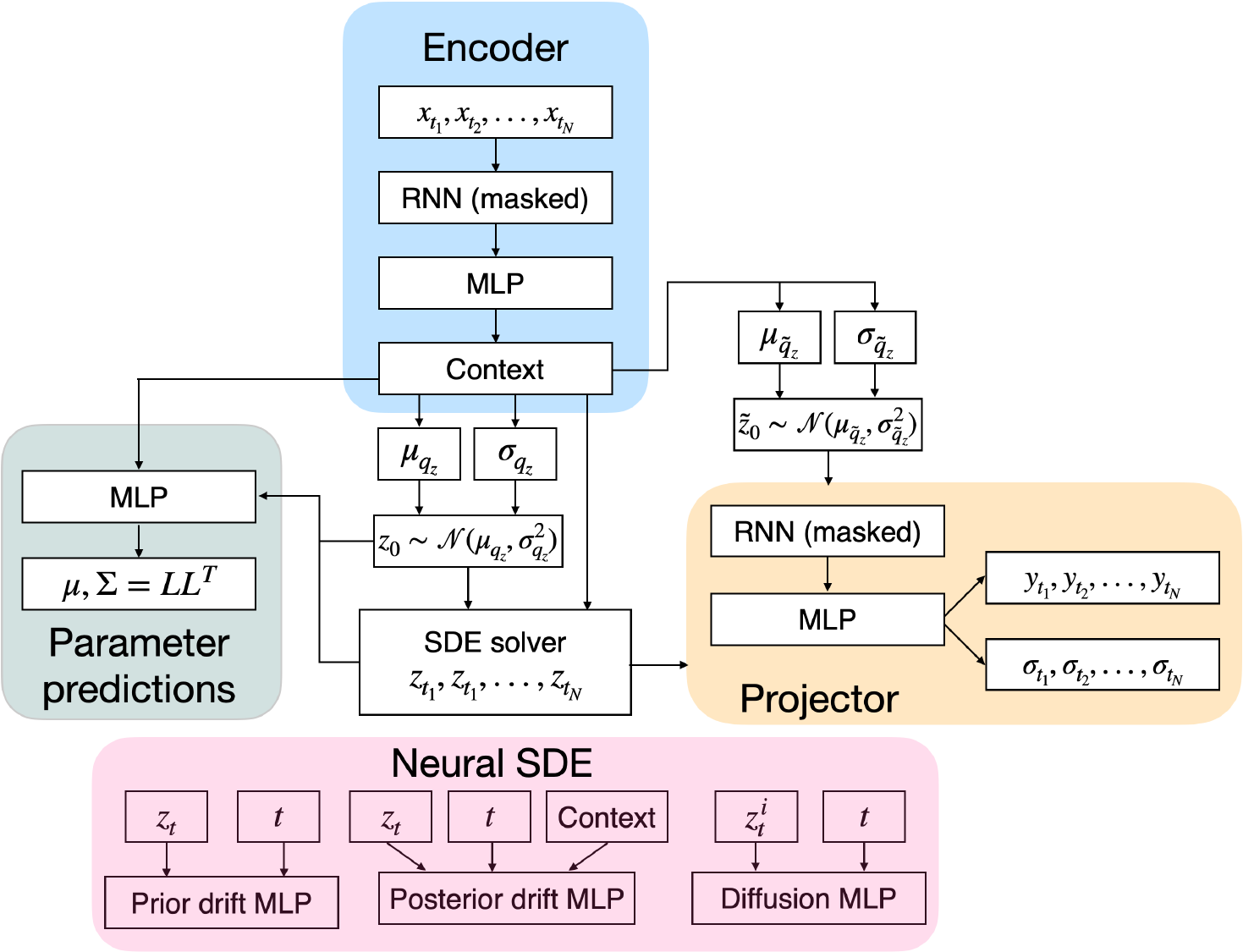}
    \caption{Model used for training the latent SDE by Fagin}
    \label{fig:faginmodel}
\end{figure}

The model has three main components (see \ref{fig:faginmodel}), and generates about $9\times 10^{5}$ total trainable parameters.

\medskip\noindent{\bf Encoder}\;
A GRU-D network \cite{Che2018} processes 12 features per time step (6 magnitudes + 6 errors) \emph{backward} in time, producing a 64-dimensional context vector $c$ that conditions the posterior drift $h_{\xi}(\cdot,\cdot;c)$. GRU-D handles missing values through a learned decay mechanism: when a band is unobserved, its hidden state decays exponentially toward a learnable mean, with the decay rate itself learned from data. The backward encoding is important: the posterior drift at time $t$ should incorporate information from \emph{future} observations (this is inference, not causal prediction).

\medskip\noindent{\bf Neural SDE decoder}\;
The latent state $z(t)\in\mbR^{8}$ evolves according to the prior and posterior SDEs as defined in \citetext{pmlr-v108-li20i}, integrated forward with Euler-Maruyama via {\tt torchsde}. The step size is set to twice the minimum observation bin width.

\medskip\noindent{\bf Parameter estimation head}\;
An MLP takes the final encoder state and predicts Gaussian posteriors (mean and variance) over the physical parameters, trained with a supervised negative log-likelihood loss alongside the ELBO.

\subsection{Training Data and Simulation}

The training set consists of $10^{5}$ simulated 10-year LSST quasar light curves with known ground-truth physical parameters. The simulations use a DRW driving signal mapped through general-relativistic accretion disk transfer functions to six-band UV/optical light curves, then sampled with realistic LSST cadences from {\tt rubin\_sim} and noise $\sg_{\rm tot}^{2} = \sg_{\rm sys}^{2} + \sg_{\rm rand}^{2}$ with $\sg_{\rm sys}=0.005$ mag.

\subsection{Training Objective}

The total loss is a weighted sum:
\smallskip
\beq\label{e:fagin_loss}
\mathcal{L} \;=\; \mathcal{L}_{\rm NLL}
\;+\; \la_{\rm ctx}\,\mathcal{L}_{\rm ctx}
\;+\; \bg_{k}\,\mathcal{L}_{\rm KL}
\;+\; \la_{\rm param}\,\mathcal{L}_{\rm param}
\eeq
\smallskip
with $\mathcal{L}_{\rm NLL}$ the Gaussian negative log-likelihood of the reconstructed light curve, $\mathcal{L}_{\rm ctx}$ a weighted MSE at observed context points, $\mathcal{L}_{\rm KL}$ the path-space KL divergence, used in \citetext{pmlr-v108-li20i}, plus the initial-state KL (with a cyclically annealed weight $\bg_{k}$ ramping from $0$ to $1$ to prevent posterior collapse), and $\mathcal{L}_{\rm param}$ a supervised NLL for the physical parameters.

The important point is that $\mathcal{L}$ has \emph{four} sources of stochasticity: the mini-batch, the Brownian realization of the posterior SDE, the decoder sampling, and the parameter-head noise. The gradient estimator is unbiased but high-variance, and is the main reason Fagin et al.\ use Adam with a small learning rate and extensive gradient clipping.

\subsection{Results and Follow-up}

The latent SDE is compared to multi-output Gaussian process regression (GPR). GPR is the standard baseline for single-object light-curve interpolation but requires per-object fitting ($\sim$minutes per object) and assumes a fixed kernel family. Fagin et al.\ show that the latent SDE (i)~produces better reconstructions in seasonal gaps (where GPR reverts to its prior mean), (ii)~processes millions of light curves in seconds after training, and (iii)~achieves good recovery of black hole mass and disk parameters directly from photometry, a task that would otherwise require spectroscopy.

A follow-up paper \cite{Fagin_2025} makes the reconstruction and parameter inference \emph{physically self-consistent} by embedding an auto-differentiable simulator of the accretion disk directly into the computational graph: a latent SDE generates the driving X-ray variability, predicted disk parameters, determine the transfer functions, and the UV/optical light curves are obtained by convolution---all differentiable end-to-end, and all trained by SGD on the composite loss.

\section{Neural SDDEs for Astronomical Time Series: Oh et al.\ (2025)}
\label{s:sdde}

\subsection{Limitation of latent SDE}

The Latent SDE framework is Markovian: the evolution of $z(t)$ depends only on the state $z(t)$ at time $t$. However, many astrophysical systems naturally violate this assumption. In quasar accretion disk reprocessing problem, a X-ray corona that itself produces variability illuminates the accretion disk, and the disk re-emits UV or optical light with wavelength-dependent time delays $\tg_{\la}\propto\la^{4/3}$ (from the thin-disk temperature profile). The optical variability at time $t$ is a reprocessed signal of the X-ray flux at time $t-\tg_{\la}$. Consequently, the system exhibits historical dependencies that violate the Markov assumption inherent in standard Latent SDEs. While it might attempt to implicitly encode past-time features within its parametric latent space $\theta$, the neural SDE model does not explicitly account for delay effects. Therefore, latent SDE, or neural latent SDE, has limitations for modeling astronomical time series data like quasar light curve.

\begin{figure}[H]
    \centering
    \includegraphics[width=0.634\linewidth]{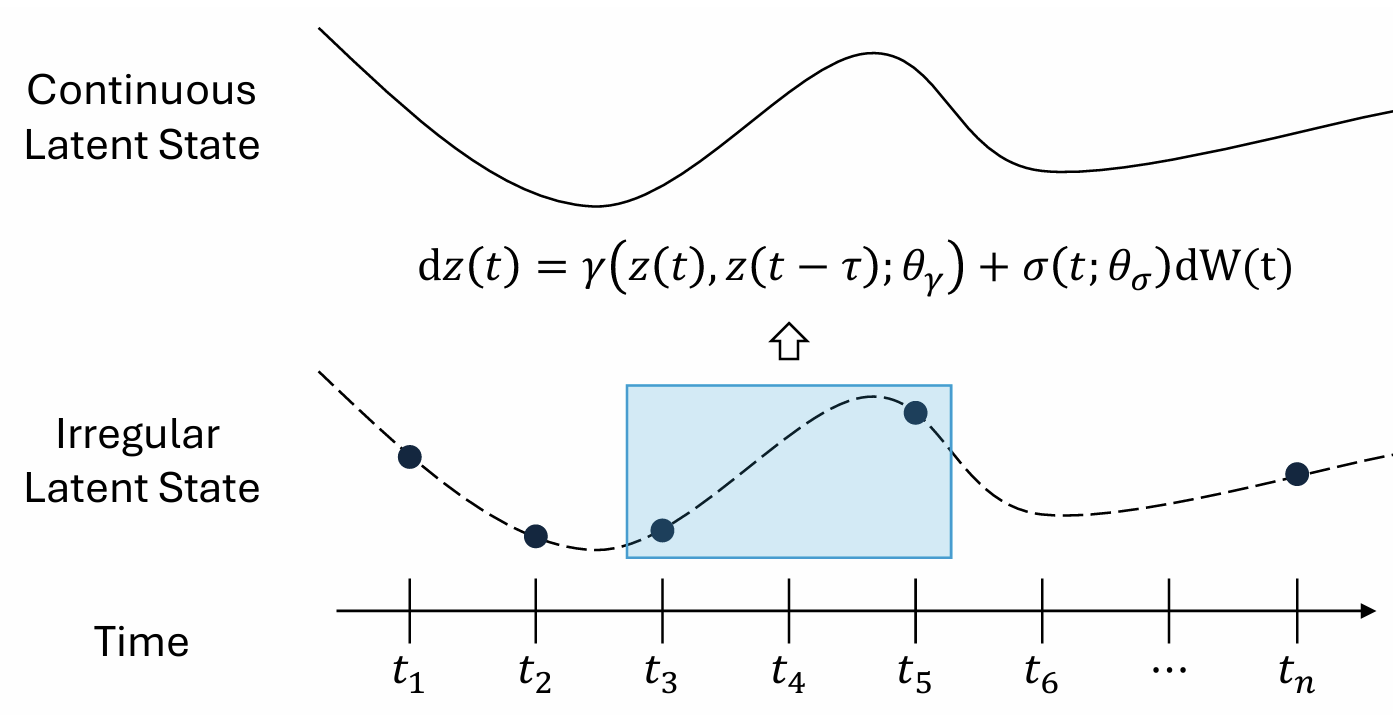}
    \caption{Neural SDDE learning window by Oh}
    \label{fig:ohdelay}
\end{figure}

\medskip\noindent Instead, the suggested model uses past observations in highlighted window (see \ref{fig:ohdelay}), and learns the dynamics from incomplete data by capturing delayed \& stochastic dependencies.

\subsection{From Neural SDE to Neural SDDE}

Neural SDE is usually written in the form:
\smallskip
\beq\label{e:sde}
\mathrm{d} z(t) \;=\; \ga \bigl(z(t);\, \thg_\ga)\,\mathrm{d} t
\;+\; \sg(z(t); \, \thg_\sg)\,\mathrm{d} W(t), \qquad t\ge 0
\eeq
\smallskip
where the drift $\ga$ governs the deterministic evolution, and diffusion $\sg$ governs stochasticity. Both of $\ga$, $\sg$ are neural networks parameterized by $\thg_\ga$ and $\thg_\sg$. By explicitly considering delayed effects, \citetext{10.1145/3746252.3760805} extend this system to define the Neural Stochastic Delay Differential Equations (Neural SDDEs):
\smallskip
\beq\label{e:sdde}
\mathrm{d} z(t) \;=\; \ga \bigl(z(t), z(t - \tau);\, \thg_\ga)\,\mathrm{d} t
\;+\; \sg(t; \, \thg_\sg)\,\mathrm{d} W(t), \qquad t\ge 0
\eeq
\smallskip
with an initial segment $z(t)=\fg(t)$ for $t\in[-\tg,0]$, where $\tg>0$ is a fixed delay hyperparameter. Now, the drift $\ga_{\thg}$ depends on both $z(t)$ and the delayed state $z(t-\tg)$. This change allows the model to capture long-range dependencies in the process.

\medskip\noindent{\bf Function-space Markov property}\;
Although~\refeq{sdde} is non-Markovian as a process on $\mbR^{d}$, it becomes Markovian when $z_{t}(\cdot) := z(t+\cdot)$ is considered as an element of the function space $C := C([-\tg,0];\mbR^{d})$. $z_{t}(\cdot)$ is called a $C$-valued Markov process \cite{mao2007stochastic}.

\medskip\noindent{\bf Reconstruction Property}\;
SDDEs exhibit a unique reconstruction property, which implies that the system’s initial history function can be recovered using only a future segment of the solution path, notably without requiring knowledge of the specific noise trajectory. This characteristic enables the adjoint method for memory-efficient backpropagation

\medskip\noindent{\bf Augmented state with controlled paths}\;
To inject information from irregular observations into the continuous-time dynamics, Oh et al.\ borrow from Neural CDEs \cite{10.5555/3495724.3496286} and interpolate the raw observations to a continuous path $X(t)$. The state is then augmented as $\bar z(t) = \zg(t, z(t), X(t); \thg_{\zg})$, where $\zg$ is a neural network, $z(t)$ is the current latent state, and $X(t)$ is a controlled path.

\subsection{Adjoint Method for SDDEs}

Computing $\partial\mathcal{L}/\partial\thg$ through~\refeq{sdde} requires a delay-aware version of the stochastic adjoint. The key complication is that the delay $\tg$ in the forward pass becomes an \emph{advance} in the backward pass: the adjoint $a(t) = \partial\mathcal{L}/\partial z(t)$ receives contributions both from the current-time derivative $\partial\ga/\partial z(t)$ and from a future time $t+\tg$ (where $z(t)$ appears as the delayed argument of the drift). Oh et al.\ implement this by partitioning the time interval into segments of length $\le \tg$ and solving the backward SDE on each segment, carrying forward the necessary future adjoint values.

\subsection{Experiments and Results}

The experiments use the ELAsTiCC and PLAsTiCC datasets (simulated LSST-like light curves across many object classes). Four scenarios are considered: standard supervised classification, classification with $50\%$ missing labels, novelty detection, and joint classification and novelty detection with missing labels. Baselines include GRU-D, Neural ODE, Neural SDE, Neural CDE, ODE-RNN, and Neural LSDE. The Neural SDDE consistently achieves the highest accuracy and weighted F$_1$, with the largest margins in the missing-label and novelty-detection settings. Sensitivity to $\tg$ is mild.

\section{Delayed-Memory Gradient Descent}
\label{s:sdgd}

\subsection{Mathematical Definitions}

\subsubsection{Neural SDDEs in Langevin form}

A special class of Neural SDDEs is the Neural Langevin-type Stochastic Delay Differential Equation (Neural LSDDE), inspired from \citetext{10.1145/3746252.3760805} and Langevin dynamics. Let $X(t)\in\mathbb{R}^{d_{X}}$ be the dynamic state at current time $t$, then the Neural Stochastic Delay Differential Equation (Neural SDDE) can be written as:
\smallskip
\beq\label{e:n_sdde}
\mathrm{d} X(t) = \gamma(X(t), X(t - \tau); \theta_\gamma) \mathrm{d} t + \sigma(t, X(t), X(t - \tau); \theta_\sigma) \mathrm{d} W(t)
\eeq
\smallskip
where $\gamma: \mathbb{R}^{d_X} \times \mathbb{R}^{d_X} \rightarrow \mathbb{R}^{d_X}$, and $\sigma: \mathbb{R}_+ \times \mathbb{R}^{d_X} \rightarrow \mathbb{R}^{m}$ are drift and diffusion functions, and $\gamma$ is a latent neural network parameterized by $\theta_\gamma$, and $\theta_\gamma$ is learned directly from available data (Similar for $\theta_\sigma$).

\medskip\noindent In ~\refeq{n_sdde}, $\tau > 0$ is a fixed delay, and $X(t-\tau)$ denote the dynamic state at past time $t-\tau$. Because of this delay, the system can not be initialized by a single point $X(0)$ but with a full history segment of the trajectory:
\smallskip
\beq\label{e:phi}
X(t) = \phi(t), \text{ } \forall t \in [-\tau, 0], \qquad \text{ and the function } \phi: [-\tau, 0] \rightarrow \mathbb{R}^{d_X}
\eeq
\smallskip
Equation~\refeq{n_sdde} explicitly integrates past states $X(t - \tau)$, allowing the model to capture memory effects. Also, $\tau$ serves as a lookback window for the model to learn temporal dependencies.

\subsubsection{Stochastic Delay Gradient Flow}

The continuous time deterministic Delayed-Memory Gradient Flow can be written as:
\smallskip
\beq\label{e:flow1}
\mathrm{d} X_t = -(\nabla f(X_t) + \lambda(X_t - X_{t - \tau})) \mathrm{d} t
\eeq
\smallskip
with the gradient-flow term $-\nabla f(X_t)$ and the delayed-memory feedback term $-\lambda(X_t - X_{t - \tau})$. The parameter $\lambda$ here is a delay-coupling strength parameter, determining how strong the delay-memory effect is.

\medskip\noindent We can also create an extension of~\refeq{flow1} by introducing stochasticity, capturing system uncertainty and robustness to noise:
\smallskip
\beq\label{e:flow2}
\mathrm{d} X_t = -(\nabla f(X_t) + \lambda(X_t - X_{t - \tau})) \mathrm{d} t + \sqrt{\eta} \Sigma(X_t, X_{t - \tau})^{1/2} \mathrm{d} W_t
\eeq
\smallskip
with $W_t$ a $d$-dimensional Brownian motion, $\Sigma$ representing the diffusion matrix, and $\eta$ representing the learning rate. Notice that~\refeq{flow2} is a special case of~\refeq{n_sdde}, with $\gamma = -(\nabla f(X_t) + \lambda(X_t - X_{t - \tau}))$, and $\sigma = \sqrt{\eta} \Sigma(X_t, X_{t - \tau})^{1/2}$.

\subsection{Discrete-time Algorithms}

\subsubsection{Delayed-Memory Stochastic Gradient Descent}

We start with the basic discrete-time gradient algorithm by considering non-stochastic and non-latent-neural case, by transform~\refeq{flow1} into a iterated version. Let $m \in \mathbb{N}$ be a delay parameter representing how many steps furthermost does the algorithm consider, and let $\eta >0$ be the learning rate, $\lambda \ge 0$ be the coupling parameter (also called memory strength parameter), we can define the discrete history segment as:
\[
x_{-m}, x_{-m+1}, \dots, x_{-1}, x_{0} \in \mathbb{R}^{d_X} \text{, then:}
\]
\beq\label{e:dmgd}
x_{k+1} = x_k - \eta \nabla f(x_k) + \eta \lambda (x_{k - m} - x_k)
\eeq
\smallskip
is the basic Delayed-Memory Gradient Descent (DMGD) iteration formula.

\medskip\noindent In real optimization problems, the Gradient Descent algorithm is usually outperformed in practice by Stochastc Gradient Descent (SGD), which generally uses a mini-batch subset of data instead of the entire dataset to determine the gradient at each iteration $k$. The advantages of SGD are faster updates, reduced memory capacity, and more likely to escape local minima or saddle points. Therefore, we can design a counterpart of SGD for the basic DMGD:
\smallskip
\beq\label{e:dmsgd}
x_{k+1} = x_k - \eta \nabla \widehat{f(x_k)} + \eta \lambda (x_{k - m} - x_k)
\eeq
\smallskip
where $\nabla \widehat{f(x_k)} = (1/b)[\sum\limits_{i\in B_k} \nabla f_i(x_k)]$, and $B_k$ is a mini-batch with $|B_k| = b$ randomly selected from the entire dataset at each $k$.

\medskip\noindent Now, focusing on the term $\lambda (x_{k - m} - x_k)$, which is called the delay feedback term, the memory strength parameter $\lambda \ge 0$ determines how strongly the past state affects the current dynamics. Overall, this term's role can be described as "besides moving downhill, the system also feels a restoring/pulling force toward its past state, or it penalizes moving too far from the past."

\subsubsection{Continuous-Delayed-Memory SGD}

In many astronomical systems' data, like irregularly sampled quasar light curves and quasar visibility, from different distances and different radii, bands, and reprocessing regions, the delays can not ideally be concentrated at a single lag $\tau$. Instead, the present response may depend on a distribution of past lags, reflecting propagation, scattering, and reprocessing across multiple spatial or physical regions \cite{Cackettetal2007}. Thus, a single-delay term $X_t - X_{t-\tau}$ may be too restrictive once the basic delayed-memory mechanism has been understood. The integral
\[
\int_{0}^{\tau} \Lambda(s)\bigl(X_t - X_{t-s}\bigr)\, \mathrm{d}s
\]
is used to model a continuum of delays instead of a single delay $\tau$, and $\Lambda(s)$ is a memory kernel function representing how much influence size-$s$ delay has, where $s \in [0, \tau]$. $\Lambda$ describes how strongly past states contributes to the present update through a distributed memory mechanism. In this sense, $\Lambda(s)$ serves as a distributed-delay analogue of the coupling parameter $\lambda(s)$ in the Basic Delayed-Memory model. Now, if only focusing on the deterministic part of the continuous-time flow:
\smallskip
\beq\label{e:Cdmgd}
\mathrm{d} X_t = - (\nabla f(X_t) + \int_{0}^{\tau} \Lambda(s)\bigl(X_t - X_{t-s}\bigr)\, \mathrm{d}s) \mathrm{d} t
\eeq
\smallskip
Therefore, for a discrete-time algorithm, let $\{w_j\}_{j=1}^m$ be the set of nonnegative weights approximating the continuous kernel, then~\refeq{Cdmgd} can be written as an iteration map:
\smallskip
\beq\label{e:Cdmgd2}
x_{k+1} = x_k - \eta \nabla f(x_k) + \eta \sum_{j = 1}^m w_j(x_{k - j} - x_k)
\eeq
\smallskip
Besides, the Basic Delayed-Memory model can be viewed as a limiting-special case of the Continuous-Delayed-Memory model with $\Lambda(s)$:if \(\Lambda(s)=\lambda\,\delta(s-\tau_0)\), where \(\delta\) is the Dirac-delta function at \(\tau_0\). Thus,~\refeq{Cdmgd},~\refeq{Cdmgd2} can be simplified as 
\[
\mathrm{d} X_t = -(\nabla f(X_t)+\lambda(X_t-X_{t-\tau_0})) \mathrm{d} t, \qquad \text{ } x_{k+1} = x_k - \eta \nabla f(x_k) + \eta \lambda (x_{k - m} - x_k)
\]
Hence, the Continuous-Delayed-Memory model is a strictly generalized version of the previous model in \textbf{Section 4.2.1}. Finally, by considering the stochastic part, we add the diffusion term to~\refeq{Cdmgd}:
\smallskip
\beq\label{e:Cdmsgd}
\mathrm{d} X_t = - (\nabla f(X_t) + \int_{0}^{\tau} \Lambda(s)\bigl(X_t - X_{t-s}\bigr)\, \mathrm{d}s) \mathrm{d} t + \sqrt{\eta} \Sigma(X_t, \{X_{t - s})^{1/2}\}_{s\in [0, \tau]} \mathrm{d} W_t
\eeq

\subsection{Solution Analysis about Stochastic Delayed Differential Equation (SDDE)}

In this section, the goal is to mathematically analyze the assumptions under which the SDDE that is a continuous-time approximation to our designed discrete algorithms has existed, unique, and global solutions, specifically~\refeq{flow2},~\refeq{Cdmsgd}. Because every SDDE is a special, often simpler, case of a Stochastic Functional Differential Equation (SFDE), and there have many existed theories about assumptions required for the existence and uniqueness of solutions to SFDE, like in \citetext{Songetal2013}, all findings are inherited from \cite{Songetal2013} but we create the variants only specific to SDDEs~\refeq{flow2}and~\refeq{Cdmsgd}.

\subsubsection{Necessary condition for SDDE~\refeq{flow2}}

To prove that this exact model has a unique, global (non-exploding) solution using the Khasminskii-type approach (bypassing strict linear growth), the following two specific mathematical conditions must hold.

\medskip\noindent \textbf{Condition 1: Local Lipschitz Continuity (Guarantees Uniqueness)}

\medskip\noindent Let $X_t$ be the current state and $X_{t - \tau}$ be the delayed state, for any local region bounded by radius $R$, $\exists \text{ }K_R \in \mathbb{R}$, s.t.
\[
\begin{aligned}
& \forall \text{ two pairs of states, } (X_{\displaystyle t_1}, X_{\displaystyle t_1 - \tau_1}), (X_{\displaystyle t_2}, X_{\displaystyle t_2 - \tau_2}),\\
& \left|-\nabla f(X_{\displaystyle t_1}) - \lambda(X_{\displaystyle t_1} - X_{\displaystyle t_1 - \tau_1}) - (-\nabla f(X_{\displaystyle t_2}) - \lambda(X_{\displaystyle t_2} - X_{\displaystyle t_2 - \tau_2}))\right|^2\\
\vee & \left|\sqrt{\eta}\Sigma(X_{\displaystyle t_1}, X_{\displaystyle t_1 - \tau_1})^{1/2} - \sqrt{\eta}\Sigma(X_{\displaystyle t_2}, X_{\displaystyle t_2 - \tau_2})^{1/2}\right|^2\\
& \le K_R(\left|X_{\displaystyle t_1} - X_{\displaystyle t_2}\right|^2 + \left|X_{\displaystyle t_1 - \tau_1} - X_{\displaystyle t_2 - \tau_2}\right|^2)
\end{aligned}
\]
\smallskip
Because the delay term $\lambda(X_t - X_{t - \tau})$ is strictly linear, it is automatically globally Lipschitz. Thus, whether Lipschitz Continuity satisfied or not is entirely determined on objective function $f$ and noise $\Sigma$.

\medskip\noindent Because the drift function, which contains the gradient $\nabla f(X_t)$, should be locally Lipschitz continuous, for any local bounded region (a compact set $E$), there must exist a constant $L_E > 0$ such that for any $X_{t_1}, X_{t_2} \in E$, $\|\nabla f(X_{t_1}) - \nabla f(X_{t_2})\| \le L_E \|X_{t_1} - X_{t_2}\|$.

\medskip\noindent Suppose $f(X_t) \in C^2$, then its second derivative, the Hessian matrix $\nabla^2 f(X_t)$, exists and is continuous everywhere.

\medskip\noindent By Multidimensional Mean Value Theorem, the difference between the gradients at two stages $X_{t_1}$ and $X_{t_2}$ can be expressed as an integral of the Hessian along the straight line path between them:
\[
\nabla f(X_{t_1}) - \nabla f(X_{t_2}) = \left( \int_0^1 \nabla^2 f\big(X_{t_2} + t(X_{t_1}-X_{t_2})\big) dt \right) (X_{t_1} - X_{t_2})
\]
\medskip\noindent Normalize both sides and by Submultiplicativity of Induced Matrix Norm, Triangle Inequality for Integrals:
\smallskip
\beq\label{e:pf1}
\|\nabla f(X_{t_1}) - \nabla f(X_{t_2})\| \le \left( \int_0^1 \big\|\nabla^2 f\big(X_{t_2} + t(X_{t_1}-X_{t_2})\big)\big\| dt \right) \|X_{t_1} - X_{t_2}\|
\eeq
\smallskip
Because $f(X_t) \in C^2$, its Hessian $\nabla^2 f(X_t)$ is still a continuous function. By Extreme value theorem, $\nabla^2 f(X_t)$ evaluated on a closed, bounded region $E$ must have a finite maximum value. So, the normalized maximum bound value $L_E = \max_{Z_t \in E} \|\nabla^2 f(Z_t)\|$ exists.

\medskip\noindent Then, replace the integral in~\refeq{pf1} with this maximum bound value, we have
\[
\|\nabla f(X_{t_1}) - \nabla f(X_{t_2})\| \le \left( \int_0^1 L_E dt \right) \|X_{t_1} - X_{t_2}\| = L_E \|X_{t_1} - X_{t_2}\|
\]
Similarly, the diffusion term $\sigma(X_t, X_{t-\tau}) = \sqrt{\eta}\Sigma(X_t, X_{t-\tau})^{1/2}$ should also be locally Lipschitz:
\smallskip
\beq\label{e:pf2}
\|\sigma(X_{t_1}, X_{t_1-\tau_1}) - \sigma(X_{t_2}, X_{t_2-\tau_2})\| \le M_E \big(\|X_{t_1} - X_{t_2}\| + \|X_{t_1-\tau_1} - X_{t_2-\tau_2}\|\big)
\eeq
\smallskip
Now, suppose that $\sigma(X_t, X_{t-\tau}) \in C^1$, then its Jacobian matrix $J_\sigma$ is continuous. Then, on any bounded compact region $E$, this continuous Jacobian will attain a finite maximum norm $M_E = \max_{X_t \in E} \|J_\sigma(X_t)\|$.

\medskip\noindent By Extreme value theorem, this result guarantees~\refeq{pf2}.

\medskip\noindent Therefore, $\forall X_t \in E, \big(f(X_t) \in C^2 \big) \land \big( \sigma(X_t, X_{t-\tau}) \in C^1 \big) \implies \big( L_E = K_R$ is the Lipschitz constant$\big)$.

\medskip\noindent \textit{(\textbf{Note:} In stochastic analysis, authors often blanket-assume $C^2$ smoothness for all coefficients---both drift and diffusion. This is because $C^2$ continuity is strictly required later to apply \textbf{It\^o's Lemma}. It\^o's Lemma fundamentally relies on a second-order Taylor expansion to calculate the Infinitesimal Generator $\mathcal{L}V(x, y)$ that is used in the Khasminskii theorem).}

\medskip\noindent \textbf{Condition 2: The Khasminskii Lyapunov Bound (Guarantees Global Existence)}

\medskip\noindent By \textbf{Theorem 2.6} in \cite{Songetal2013},

\noindent we construct a non-negative Lyapunov function (also called "containment" function) $V(X_t)$. Specifically, for our SDDE model, the Infinitesimal Generator operator $\mathcal{L}$ acting on a twice-differentiable function $V(X_t)$ is defined using It\^o's Lemma as:
\[
\mathcal{L}V(X_t, X_{t-\tau}) = \nabla V(X_t)^\top \gamma(X_t, X_{t-\tau}) + \frac{1}{2}\text{Tr}\Big[\sigma(X_t, X_{t-\tau})^\top \nabla^2 V(X_t) \sigma(X_t, X_{t-\tau})\Big]
\]
where $\gamma$ is the drift vector and $\sigma$ is the diffusion matrix. Substituting our specific drift $\gamma = -\big(\nabla f(X_t) + \lambda(X_t - X_{t-\tau})\big)$ and diffusion $\sigma = \sqrt{\eta}\Sigma(X_t, X_{t-\tau})^{1/2}$:
\smallskip
\begin{equation}\label{e:vsub}
\begin{split}
\mathcal{L}V(X_t, X_{t-\tau}) = & -\nabla V(X_t)^\top \big(\nabla f(X_t) + \lambda(X_t - X_{t-\tau})\big) \\
& + \frac{\eta}{2}\text{Tr}\Big[\Sigma(X_t, X_{t-\tau})^{1/2} \nabla^2 V(X_t) \Sigma(X_t, X_{t-\tau})^{1/2}\Big]
\end{split}
\end{equation}
\smallskip

\medskip\noindent To make this mathematically actionable, let us test the standard quadratic Lyapunov containment function: $V(X_t) = \|X_t\|^2$. This implies its gradient is $\nabla V(X_t) = 2X_t$ and its Hessian is $\nabla^2 V(X_t) = 2I$ (where $I$ is the identity matrix). 

\medskip\noindent Plugging these exact derivatives into our generator~\refeq{vsub} yields:
\smallskip
\begin{equation}\label{e:vquad}
\begin{split}
\mathcal{L}V(X_t, X_{t-\tau}) = & 2X_t^\top \Big(-\nabla f(X_t) - \lambda X_t + \lambda X_{t-\tau}\Big) \\ & + \frac{\eta}{2}\text{Tr}\Big[\Sigma(X_t, X_{t-\tau})^{1/2} (2I) \Sigma(X_t, X_{t-\tau})^{1/2}\Big]
\end{split}
\end{equation}
\smallskip
Since matrix trace satisfies the cyclic property $\text{Tr}[A B] = \text{Tr}[B A]$, the diffusion term simplifies perfectly to $\eta\text{Tr}[\Sigma(X_t, X_{t-\tau})]$. Expanding the inner product for the drift term, we obtain:
\smallskip
\beq\label{e:vexpand}
\mathcal{L}V(X_t, X_{t-\tau}) = -2X_t^\top \nabla f(X_t) - 2\lambda\|X_t\|^2 + 2\lambda(X_t^\top X_{t-\tau}) + \eta\text{Tr}\Big[\Sigma(X_t, X_{t-\tau})\Big]
\eeq
\smallskip
\medskip\noindent By applying Young's inequality, which states that $2(X_t^\top X_{t-\tau}) \le \|X_t\|^2 + \|X_{t-\tau}\|^2$, we can establish a strict upper bound for the cross-term. Substituting this inequality into~\refeq{vexpand}:
\[
\mathcal{L}V(X_t, X_{t-\tau}) \le -2X_t^\top \nabla f(X_t) - \lambda\|X_t\|^2 + \lambda\|X_{t-\tau}\|^2 + \eta\text{Tr}\Big[\Sigma(X_t, X_{t-\tau})\Big]
\]
\medskip\noindent The Khasminskii condition dictates that for the system state to not explode to infinity, there must exist a bounding constant $c > 0$ such that the generator is constrained by the current and delayed size of $V$:
\[
\mathcal{L}V(X_t, X_{t-\tau}) \le c\big(1 + \|X_t\|^2 + \|X_{t-\tau}\|^2\big)
\]
\medskip\noindent Therefore, for the DMGD SDDE to admit a unique, global, non-exploding solution, the objective function $f(X_t)$ and the diffusion matrix $\Sigma$ must rigorously satisfy the following inequality for some constant $c > 0$:
\smallskip
\beq\label{e:vbound2}
-2X_t^\top \nabla f(X_t) + \eta\text{Tr}\Big[\Sigma(X_t, X_{t-\tau})\Big] \le c(1 + \|X_t\|^2) + (c - \lambda)\|X_{t-\tau}\|^2 + \lambda\|X_t\|^2
\eeq
\medskip

\noindent In summary,~\refeq{vbound2} and $\forall X_t \in E, \big(f(X_t) \in C^2 \big) \land \big( \sigma(X_t, X_{t-\tau}) \in C^1 \big)$ should both satisfied to garantee unique, global solution for~\refeq{flow2}.

\subsubsection{Necessary condition for SDDE~\refeq{Cdmsgd}}

Following the Khasminskii-type approach as done in \textbf{Section 4.3.1}, and adapting it specifically for the integral delay functional, the model~\refeq{Cdmsgd} possesses a unique, global solution if the following conditions holds:

\medskip\noindent \textbf{Condition 1: Local Lipschitz Continuity (Guarantees Uniqueness)}

\medskip\noindent For any local compact region bounded by radius $R$, denoted as $E$, let $(X_{t_1}, \{X_{t_1-s}\}_{s \in [0, \tau]})$ and $(X_{t_2}, \{X_{t_2-s}\}_{s \in [0, \tau]})$ be two state paths in $E$. To guarantee uniqueness, $\exists K_R > 0$ such that:
\smallskip
\begin{equation}\label{e:pf3}
\begin{split}
& \left\|\gamma(X_{t_1}, \{X_{t_1-s}\}) - \gamma(X_{t_2}, \{X_{t_2-s}\})\right\|^2 \vee \left\|\sigma(X_{t_1}, \{X_{t_1-s}\}) - \sigma(X_{t_2}, \{X_{t_2-s}\})\right\|^2 \\
& \le K_R \Big( \left\|X_{t_1} - X_{t_2}\right\|^2 + \sup_{s \in [0, \tau]} \left\|X_{t_1-s} - X_{t_2-s}\right\|^2 \Big)
\end{split}
\end{equation}
\smallskip

\noindent By linearity and the Triangle Inequality for Integrals, the distributed delay term satisfies:
\smallskip
\begin{equation*}
\begin{split}
& \left\| \int_0^\tau \Lambda(s)(X_{t_1} - X_{t_1-s})ds - \int_0^\tau \Lambda(s)(X_{t_2} - X_{t_2-s})ds \right\| \\
& \le \int_0^\tau \Lambda(s) \left\|X_{t_1} - X_{t_2}\right\| ds + \int_0^\tau \Lambda(s) \left\|X_{t_1-s} - X_{t_2-s}\right\| ds \\
& \le L_\Lambda \left( \left\|X_{t_1} - X_{t_2}\right\| + \sup_{s \in [0, \tau]} \left\|X_{t_1-s} - X_{t_2-s}\right\| \right)
\end{split}
\end{equation*}
\smallskip
where $L_\Lambda = \int_0^\tau \Lambda(s) ds < \infty$. Thus, the integral delay term is globally Lipschitz.

\medskip\noindent For the objective function $f$ and diffusion $\sigma$, assume $\big(f(X_t) \in C^2 \big) \land \big( \sigma(X_t, \{X_{t-s}\}) \in C^1 \big)$. By the Extreme Value Theorem on the compact set $E$, the continuous Hessian $\nabla^2 f$ and Jacobian $J_\sigma$ attain finite maximum norms:
\[
L_E = \max_{Z \in E} \|\nabla^2 f(Z)\|, \quad M_E = \max_{Z \in E} \|J_\sigma(Z)\|
\]

\medskip\noindent Applying the Multidimensional Mean Value Theorem yields:
\[
\|\nabla f(X_{t_1}) - \nabla f(X_{t_2})\| \le L_E \|X_{t_1} - X_{t_2}\|
\]
\[
\|\sigma(X_{t_1}, \{X_{t_1-s}\}) - \sigma(X_{t_2}, \{X_{t_2-s}\})\| \le M_E \Big(\|X_{t_1} - X_{t_2}\| + \sup_{s \in [0, \tau]} \|X_{t_1-s} - X_{t_2-s}\|\Big)
\]

\medskip\noindent Therefore, combining the bounded constants $L_\Lambda, L_E$, and $M_E$ algebraically guarantees the existence of a finite $K_R$, strictly satisfying~\refeq{pf3}, and thus uniqueness of the solution.

\medskip\noindent \textbf{Condition 2: The Khasminskii Lyapunov Bound (Guarantees Global Existence)}

\medskip\noindent Following \textbf{Theorem 2.6} in \citetext{Songetal2013} for Stochastic Functional Differential Equations, we apply the Infinitesimal Generator $\mathcal{L}$ to the quadratic containment function $V(X_t) = \|X_t\|^2$.

\medskip\noindent Applying It\^o's Lemma and substituting the derivatives $\nabla V(X_t) = 2X_t$ and $\nabla^2 V(X_t) = 2I$, the generator expands to:
\smallskip
\begin{equation*}
\begin{split}
\mathcal{L}V(X_t, \{X_{t-s}\}) = & 2X_t^\top \Big(-\nabla f(X_t) - \int_0^\tau \Lambda(s)(X_t - X_{t-s})ds\Big) \\
& + \frac{\eta}{2}\text{Tr}\Big[\Sigma(X_t, \{X_{t-s}\})^{1/2} (2I) \Sigma(X_t, \{X_{t-s}\})^{1/2}\Big]
\end{split}
\end{equation*}

\medskip\noindent Utilizing the cyclic property of the trace matrix and expanding the inner product into the integral, we obtain:
\smallskip
\begin{equation}\label{e:pf4}
\begin{split}
\mathcal{L}V(X_t, \{X_{t-s}\}) = & -2X_t^\top \nabla f(X_t) - 2\int_0^\tau \Lambda(s)\|X_t\|^2 ds \\
& + 2\int_0^\tau \Lambda(s)(X_t^\top X_{t-s}) ds + \eta\text{Tr}\Big[\Sigma(X_t, \{X_{t-s}\})\Big]
\end{split}
\end{equation}
\smallskip

\medskip\noindent By applying Young's inequality, $2(X_t^\top X_{t-s}) \le \|X_t\|^2 + \|X_{t-s}\|^2$, we can establish a strict upper bound for the cross-term inside the integral. Assuming the weighting function is non-negative ($\Lambda(s) \ge 0$), this yields:
\smallskip
\begin{equation*}
\begin{split}
\mathcal{L}V(X_t, \{X_{t-s}\}) \le & -2X_t^\top \nabla f(X_t) - 2\int_0^\tau \Lambda(s)\|X_t\|^2 ds \\
& + \int_0^\tau \Lambda(s)\|X_t\|^2 ds + \int_0^\tau \Lambda(s)\|X_{t-s}\|^2 ds + \eta\text{Tr}\Big[\Sigma(X_t, \{X_{t-s}\})\Big]
\end{split}
\end{equation*}
\smallskip
Which simplifies perfectly to match the structure of our previous single-delay bound:
\smallskip
\[
\mathcal{L}V(X_t, \{X_{t-s}\}) \le -2X_t^\top \nabla f(X_t) - \int_0^\tau \Lambda(s)\|X_t\|^2 ds + \int_0^\tau \Lambda(s)\|X_{t-s}\|^2 ds + \eta\text{Tr}\Big[\Sigma(X_t, \{X_{t-s}\})\Big]
\]

\medskip\noindent The general Khasminskii condition for SFDEs dictates that the generator must be bounded by a constant $c > 0$ that scales with the current state and the supremum of the delayed state path:
\[
\mathcal{L}V(X_t, \{X_{t-s}\}) \le c\Big(1 + \|X_t\|^2 + \sup_{s\in[0,\tau]}\|X_{t-s}\|^2\Big)
\]
\noindent Therefore, for the distributed-delay SDDE~\refeq{Cdmsgd} to guarantee a global solution, the objective function $f(X_t)$ and the diffusion matrix $\Sigma$ must satisfy the following inequality for some constant $c > 0$:
\smallskip
\begin{equation*}
\begin{split}
-2X_t^\top \nabla f(X_t) + \eta\text{Tr}\Big[\Sigma(X_t, \{X_{t-s}\})\Big] \le & \; c\Big(1 + \|X_t\|^2 + \sup_{s\in[0,\tau]}\|X_{t-s}\|^2\Big) \\
& + \int_0^\tau \Lambda(s)\Big(\|X_t\|^2 - \|X_{t-s}\|^2\Big) ds
\end{split}
\end{equation*}

\subsection{2D Simulation Results}

\subsubsection{Simulation on 2D Convex Quadratic Loss Landscape}

\begin{figure}[H]
    \centering
    \includegraphics[width=0.84\linewidth]{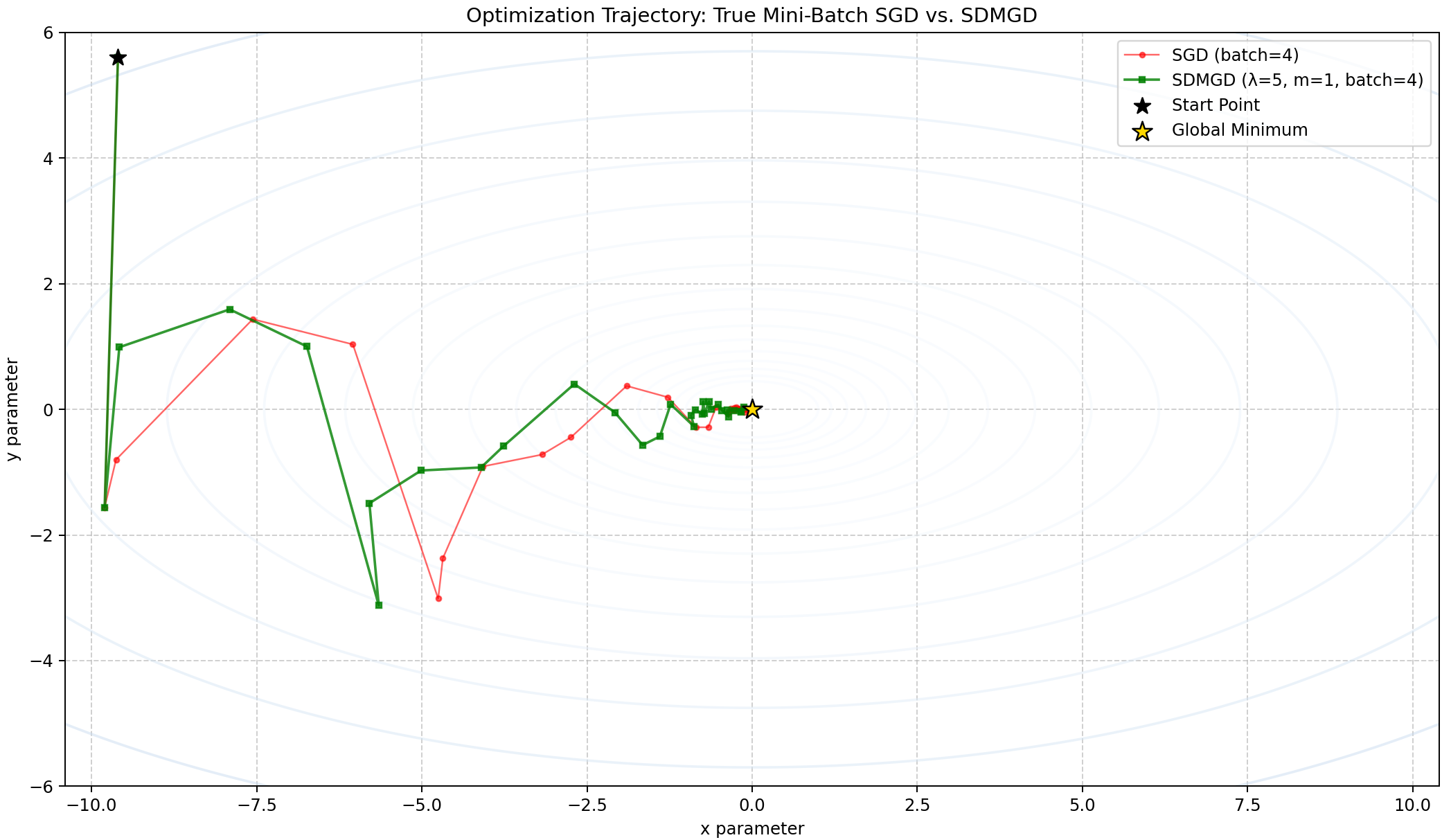}
    \includegraphics[width=0.84\linewidth]{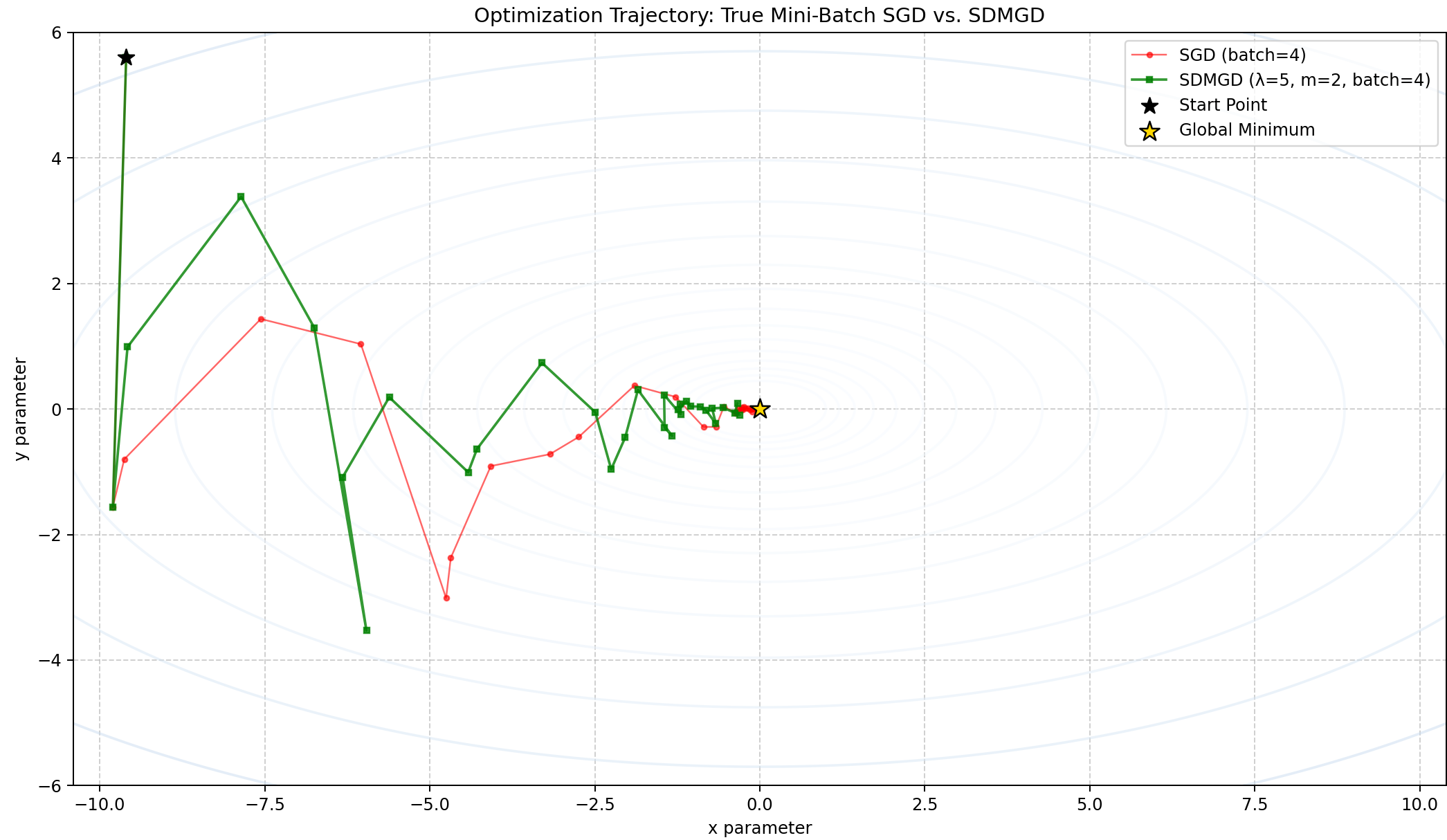}
    \includegraphics[width=0.84\linewidth]{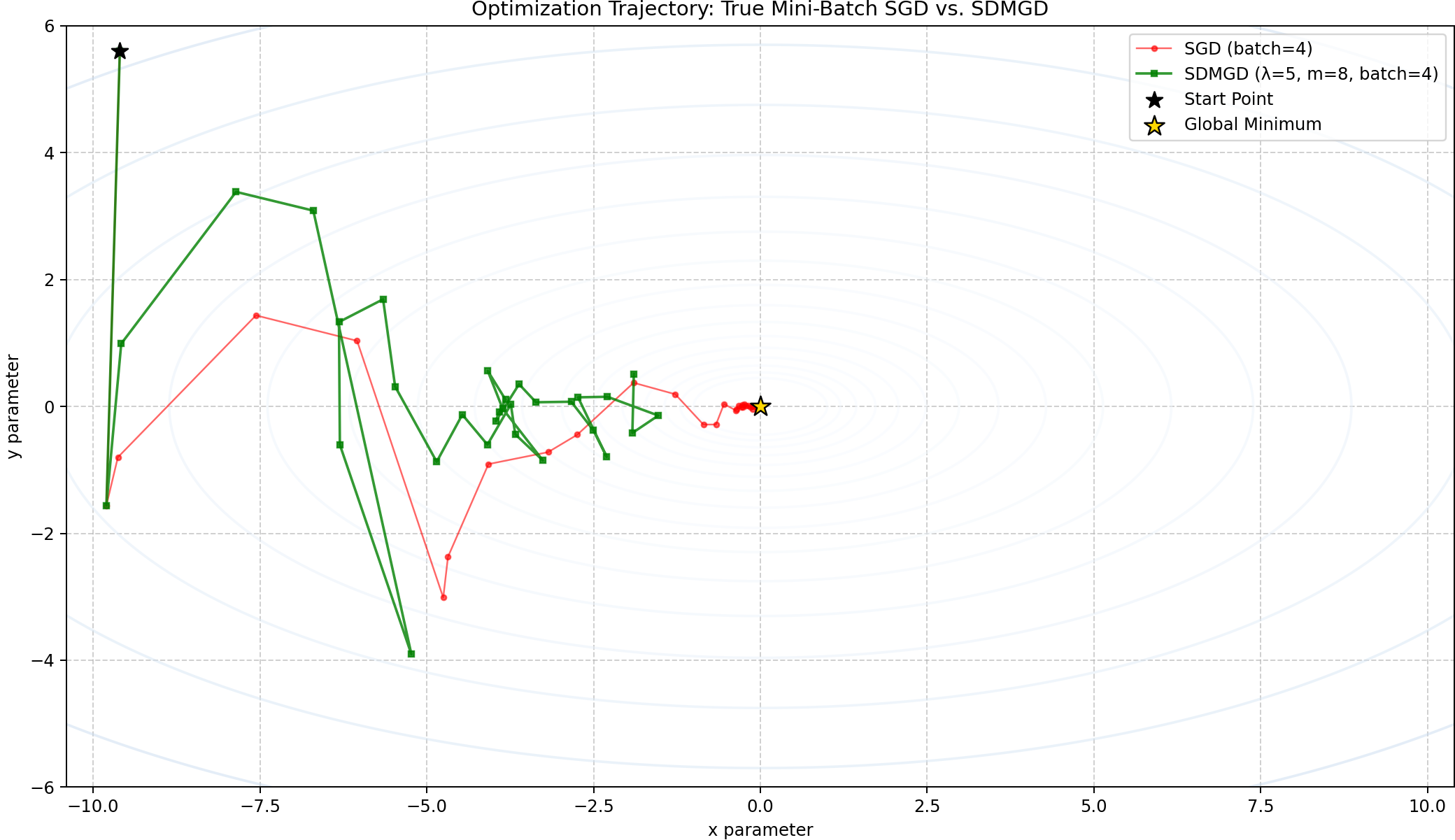}
    \caption{SDG (red) \& DMSGD (green) trajectories as $m \in \{1,2,8\}$ }
    \label{fig:simu1}
\end{figure}

We first generate a 2-dimensional synthetic dataset representing a loss landscape of $f(x, y) = x^2 + 5y^2$ with 20000 random data points, which is a typical convex quadratic loss function. Then, we control the random seed 1000, batch size $b = 4$, learning rate $\eta = 0.05$ delay-coupling strength parameter $\lambda = 5$, number of steps $k = 30$, and we gradually changes the delayed steps $m$ in the set $\{1, 2, 8\}$. Then, we run both the Vanilla SGD and the Delayed-Memory SGD on $f$ starting at initial point $(-9.6, 5.6)$. The SGD trajectory is in red and the DMSGD trajectory is in green. (See \ref{fig:simu1}, where the loss map is restricted only to $[-10.4, 10.4] \times [-6, 6]$ since the trajectories are all in this range)

\medskip\noindent The simulation shows that when keeping every other parameter constant and with small batch size, the DMSGD is likely to have more fluctuated moves on the loss landscape as the delayed steps $m$ increases, which is caused by the "long-range" phase lag forcing the optimizer to react to the geometry of a more distant region of the landscape rather than the local curvature.

\medskip\noindent Again, we use the same dataset, initial point, random seed 1000, and batch size $b = 4$, to perform both SGD and DMSGD. In the 1st round, we set the DMSGD parameters $m = 5$, $\eta = 0.004$, $\lambda = 245$, $k = 500$; in the 2nd round, we changed the DMSGD parameters $\lambda = 14$, $k = 12000$ (changing $k$ to generate enough iteration that goes to the local minimum).

\medskip\noindent See comparison plots in \ref{fig:simu1}, big $\lambda$ would lead DMSGD algorithm into divergent behavior, which helps the algorithm to explore a wider landscape in each single run. However, as $\lambda$ becomes smaller, the DMSGD trajectory is actually more smoothly-convergent to the local minimum compared to SGD's.

\begin{figure}[H]
    \centering
    \includegraphics[width=0.85\linewidth]{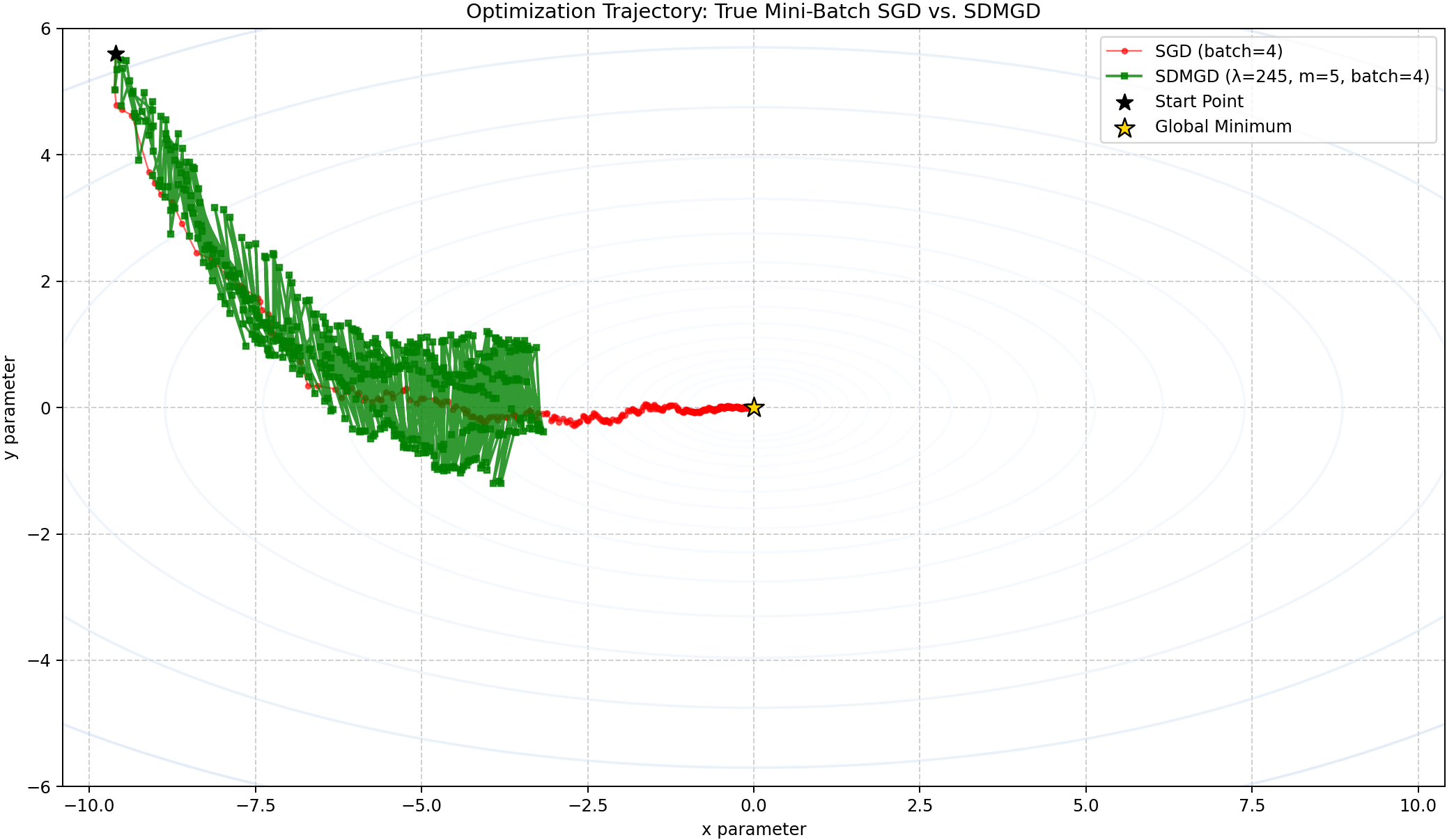}
    \includegraphics[width=0.85\linewidth]{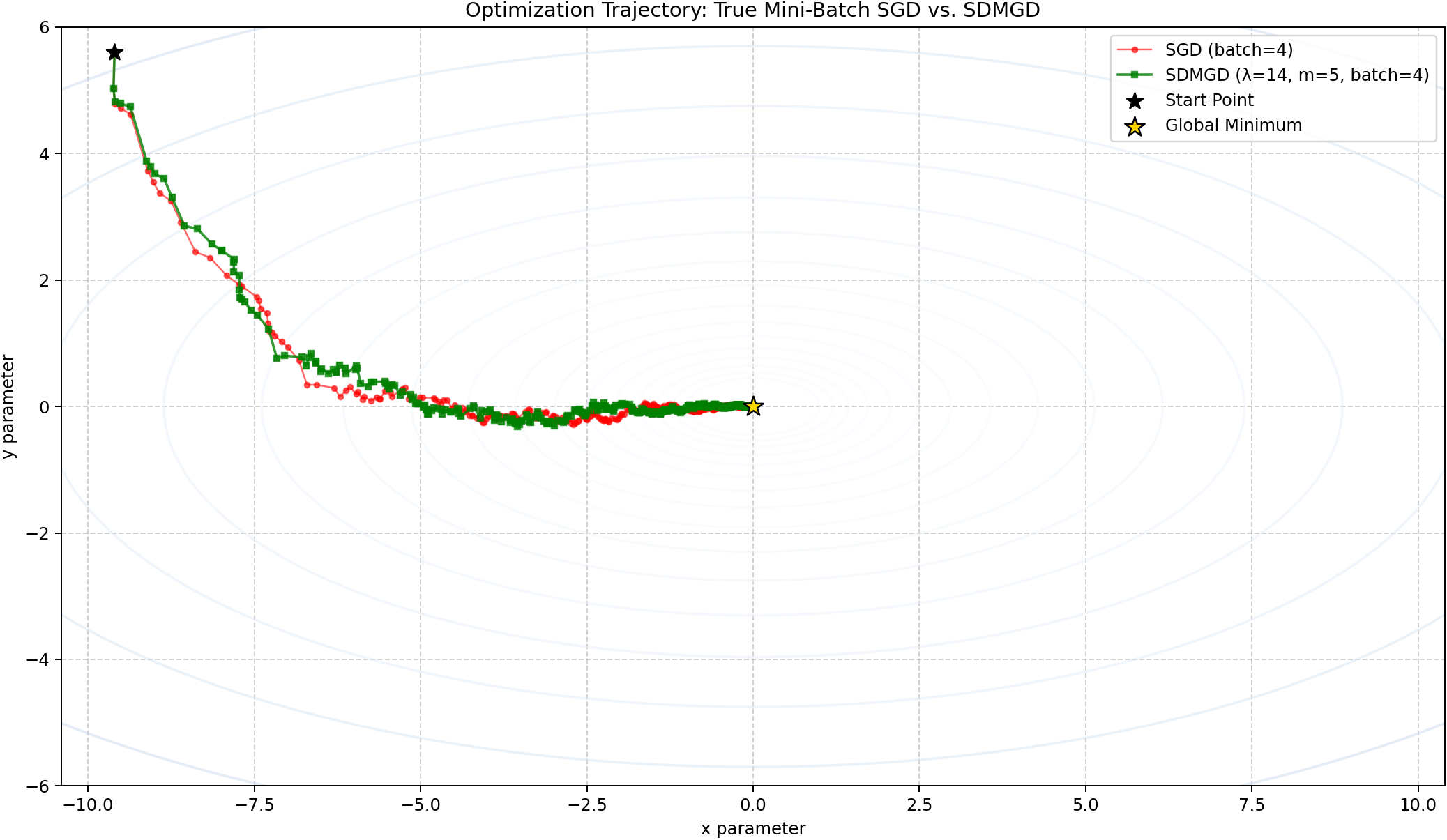}
    \caption{SDG (red) \& Continuous-DMSGD (green) trajectories as $\lambda \in \{245, 14\}$}
    \label{fig:simu2}
\end{figure}

\subsubsection{Simulation on 2D Nonconvex Rastrigin Loss Landscape}

\begin{figure}[H]
    \centering
    \includegraphics[width=0.895\linewidth]{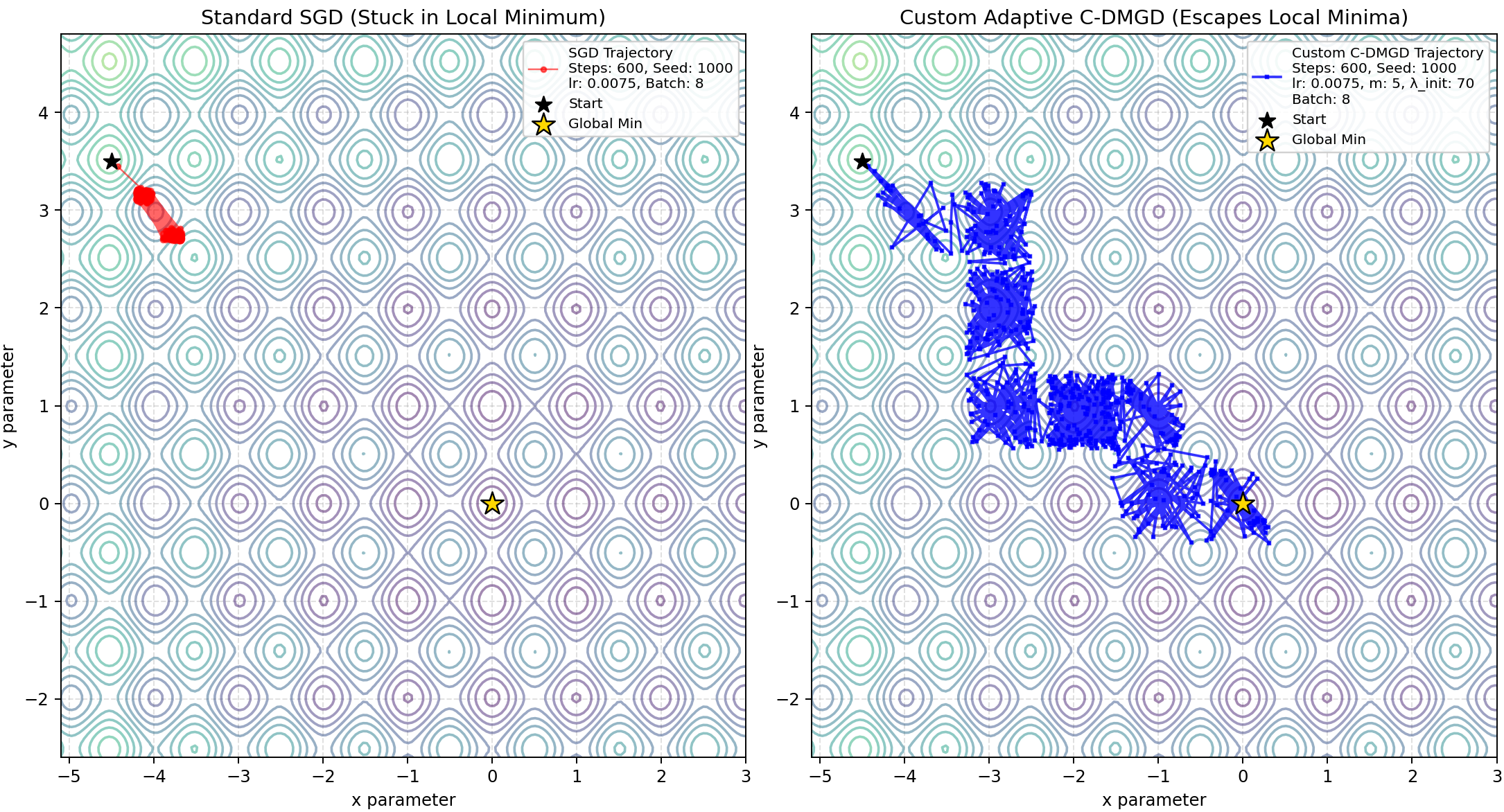}
    \caption{SDG (red) \& Continuous-DMSGD (blue) trajectories with uniform initial delay-coupling strength weight}
    \label{fig:simu3}
\end{figure}

The, we generate another 2-dimensional synthetic dataset representing a loss landscape of $f_2(\mathbf{x}) = 20 + \sum_{i=1}^{2} \left[ x_i^2 - 10\cos(2\pi x_i) \right]$ with 20000 random data points and starting point $(-4.5, 3.5)$. We use the same random seed 1000, and run Vanilla SGD and the Continuous-Delayed-Memory SGD on $f_2$. We set the batch size $b = 8$, learning rate $\eta = 0.0075$, number of steps $k = 600$, delayed steps $m = 5$ with uniform initial delay-coupling strength weight $[14, 14, 14, 14, 14]^\top$. (See \ref{fig:simu3})
\noindent In this scenario, the vanilla SGD quickly descends into the nearest local minimum and becomes permanently trapped due to its lack of momentum to overcome the walls. Conversely, the Continuous-DMSGD algorithm introduces intentional fluctuations into the path due to its pulling-back force to past locations stored in the history. These "restoring forces" allow the Continuous-DMSGD optimizer to escape various local traps and successfully navigate across the rugged terrain to converge at the global minimum in this example.

\begin{figure}[H]
    \centering
    \includegraphics[width=0.69\linewidth]{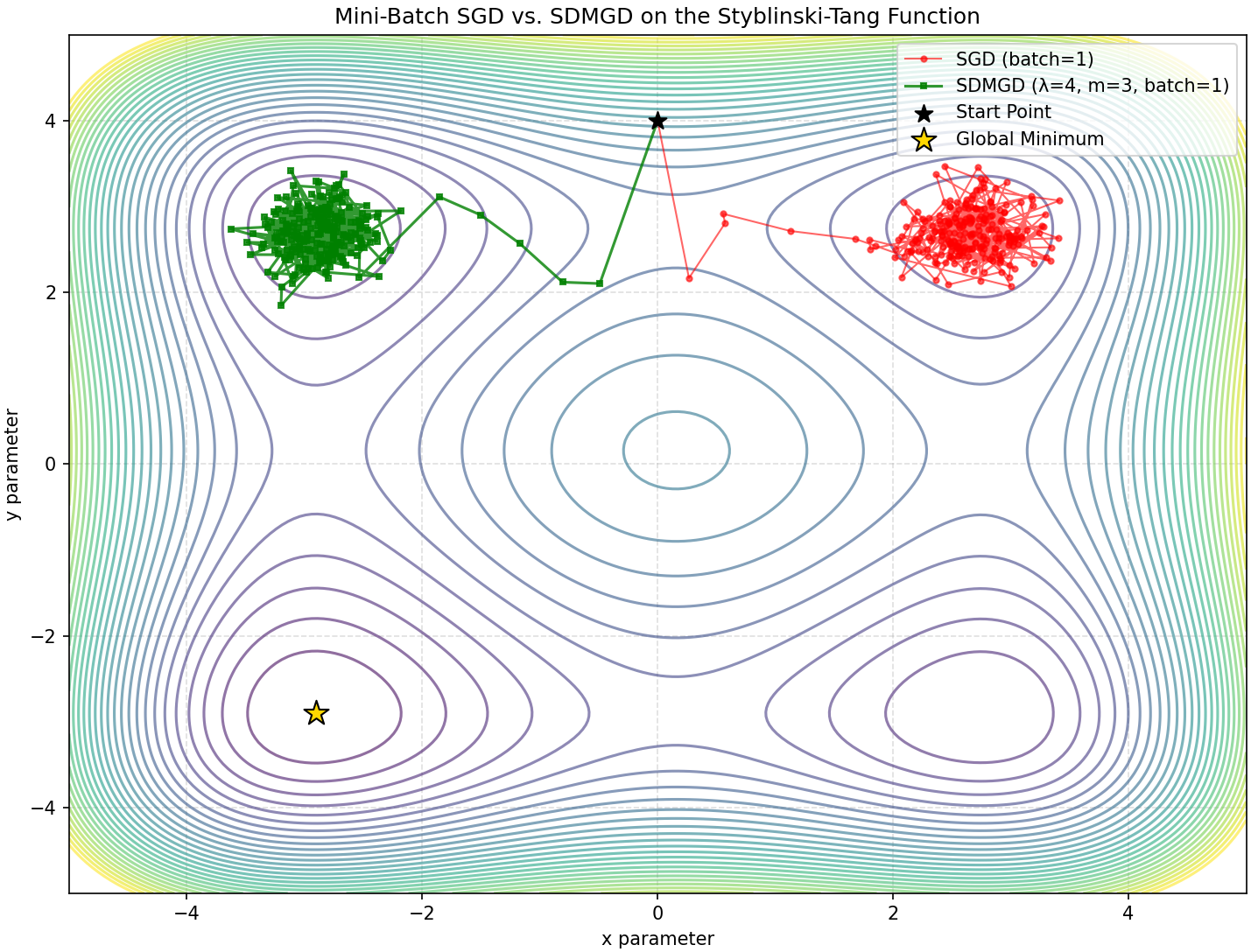}
    \caption{SDG (red) \& Continuous-DMSGD (green) trajectories finalizes into different local minimums}
    \label{fig:simu4}
\end{figure}

\medskip \noindent In another example, both SGD and Continuous-DMSDG optimizers are navigating on the 2D Styblinski-Tang Loss, represented by $f(x,y) = \frac{1}{2} \left[ (x^4 - 16x^2 + 5x) + (y^4 - 16y^2 + 5y) \right]$. (See \ref{fig:simu4}) In this certain parameter set up, the 2 optimizers finally stuck into different local minimums, which additionally supports that SGD and Continuous-DMSDG has distinct behaviors in low-dimensional space.

\subsubsection{Potential advantages of Continuous-DMSGD over Vanilla SGD}

The magnitude of the $\Lambda$ dictates a critical trade-off between convergence stability and exploration. When $\Lambda$ is small, the optimization steps becomes smaller in a local minimum region, ensuring faster convergence to the minimum. When $\Lambda$ is large, the step sizes progressively expand and allows the optimizer to explore larger regions, but it might lead to divergence. So, finding the boundary of $\Lambda$'s magnitude that determines the algorithm's different behavior is a potential future direction.
\medskip

\noindent \textbf{Observed Advantages from simulations}:
\medskip

\noindent \textbf{1}. In Highly Non-Convex Landscapes, especially those with alternating local maximums and local minimums, Continuous-DMSGD is more likely to escape some local minimum compared to Vanilla SGD through the pulling force toward its past state. This is because the pulling force from the memory term provides energy to jump over local barriers, which override the zero-gradient traps of local minima. [Increase overall magnitude of $\Lambda$]
\medskip

\noindent \textbf{2}. In very sharp local minimum region, or even non-smooth holes, Continuous-DMSGD is more likely to finally converge into lower Loss regions than Vanilla SGD by penalizing moving too far from the past. [Use very small $\Lambda$ magnitude]

 \section{From the Stochastic Adjoint to Continuous-Time Reinforcement Learning}
 \label{s:rl}

This section develops a backward Stratonovich SDE that carries the stochastic adjoint method described in \citetext{pmlr-v108-li20i}, which is the close cousin of the backward SDE that carries the costate in stochastic optimal control, and hence of the Bellman/HJB machinery of continuous-time reinforcement learning. We proceed in three steps: (i)~classical stochastic control and its BSDE adjoint; (ii)~continuous-time RL with deterministic policies, following \citetext{JMLR:v7:munos06b}; and (iii)~the exploratory (entropy-regularised) formulation of \citetext{JMLR:v21:19-144}, together with what we will call the \emph{Exploratory Backward Stratonovich SDE}.

\subsection{Controlled SDEs and the Stochastic Control Problem}

Let the state $X(t)\in\mbR^{n}$ evolve according to the controlled SDE
\beq\label{e:controlled_sde}
\mathrm{d} X(t) \;=\; b\bigl(X(t), u(t)\bigr)\,\mathrm{d} t
\;+\; \sg\bigl(X(t), u(t)\bigr)\,\mathrm{d} W(t), \qquad X(0) = x_{0}
\eeq
where $u(t)\in\mathcal{U}\subset\mbR^{k}$ is the control (action) at time $t$, chosen by the
decision maker, and $W(t)$ is an $m$-dimensional Brownian motion. The cost functional over a
horizon $T$ is
\beq\label{e:cost}
J(u) \;=\; \mathbb{E}\!\lb\int_{0}^{T} L\bigl(X(s), u(s)\bigr)\,\mathrm{d} s
+ \Fg\bigl(X(T)\bigr)\rb
\eeq
where $L$ is a running cost and $\Fg$ is a terminal cost. The value function is $V(t,x) := \inf_{u}\mathbb{E}[\int_{t}^{T}L\,\mathrm{d} s + \Fg(X(T))\mid X(t)=x]$ and, under standard regularity, satisfies the {\bf Hamilton--Jacobi--Bellman (HJB) equation}
\beq\label{e:hjb}
\partial_{t} V + \inf_{u\in\mathcal{U}}\lbr
b(x,u)^{\top}\nabla_{x}V
+ \tfrac{1}{2}\hbox{\rm tr}\bigl(\sg\sg^{\top}(x,u)\,\nabla_{x}^{2}V\bigr)
+ L(x,u)
\rbr = 0,
\qquad V(T,x) = \Fg(x)
\eeq

\subsection{The Adjoint BSDE from Pontryagin's Principle}

An alternative route to the optimal control is Pontryagin's stochastic maximum principle, which expresses optimality through a pair $(Y(t), Z(t))$ of adapted processes satisfying a \emph{backward stochastic differential equation} (BSDE) of the Pardoux--Peng type:
\beq\label{e:bsde}
\left\{
\begin{aligned}
-\mathrm{d} Y(t) &= \bigl[\,b_{x}(X,u)^{\top} Y(t)
+ \hbox{\rm tr}\!\bigl(\sg_{x}(X,u)^{\top} Z(t)\bigr)
+ L_{x}(X,u)\,\bigr]\,\mathrm{d} t - Z(t)\,\mathrm{d} W(t),\\
Y(T) &= \Fg_{x}\bigl(X(T)\bigr)
\end{aligned}
\right.
\eeq
where subscripts denote partial derivatives in $x$. The process $Y(t)$ is the costate (adjoint), and $Z(t)$ is the martingale representation term introduced to ensure the solution is adapted to the forward filtration. The optimal control maximises the Hamiltonian
\beq\label{e:hamiltonian}
H(x,u,y,z) \;=\; b(x,u)^{\top} y + \hbox{\rm tr}\!\bigl(\sg(x,u)^{\top} z\bigr) - L(x,u).
\eeq

\medskip\noindent{\it The structural parallel with the stochastic adjoint of Li et al.}\;
We find that the adjoint SDE of \citetext{pmlr-v108-li20i} is a pathwise backward SDE driven by the \emph{forward} Brownian path, whereas~\refeq{bsde} is a Pardoux--Peng BSDE whose solution must be adapted and hence involves a new martingale term $Z\,\mathrm{d} W$. But they are the same \emph{kind} of object: in both, a costate is propagated backward in time, driven in part by a Brownian motion, and used to read off gradients (or, in the control case, the Hamiltonian-maximizing action).

\subsection{Continuous-Time RL with Deterministic Policies (Munos 2006)}

\citetext{JMLR:v7:munos06b} studies policy gradient methods for deterministic feedback policies $u(t) = \pi_{\thg}(X(t))$ in continuous time. Substituting the policy into \refeq{controlled_sde} gives an SDE parameterised by $\thg$:
\beq\label{e:closed_loop_sde}
\mathrm{d} X^{\thg}(t) \;=\; b\bigl(X^{\thg}(t), \pi_{\thg}(X^{\thg}(t))\bigr)\,\mathrm{d} t
\;+\; \sg\bigl(X^{\thg}(t), \pi_{\thg}(X^{\thg}(t))\bigr)\,\mathrm{d} W(t)
\eeq
Now, compare this to the neural SDE defined in \citetext{pmlr-v108-li20i} as
\beq\label{e:neural_sde}
\mathrm{d} z(t) \;=\; f_{\thg}\bigl(z(t),t\bigr)\,\mathrm{d} t
\;+\; g_{\phi}\bigl(z(t),t\bigr)\,\mathrm{d} W(t), \qquad t\in[0,T]
\eeq
We find that these two are the similar, with parameters $\thg$ inside both the drift and the diffusion. The policy gradient $\nabla_{\thg} J(\pi_{\thg})$ can thus be computed in two equivalent ways:
\begin{enumerate}
\item via the HJB value function: $\nabla_{\thg}J = \mathbb{E}\!\bigl[\int_{0}^{T}
(\partial_{u} H)(X,u,\nabla V, \nabla^{2}V)\,\nabla_{\thg}\pi_{\thg}(X)\,\mathrm{d} t\bigr]$
\item via the stochastic adjoint of \citetext{pmlr-v108-li20i}: solve the forward SDE \refeq{closed_loop_sde}, then solve the backward Stratonovich SDE against the reward-based loss $\mathcal{L}(X^{\thg}(\cdot)) = \int L\,\mathrm{d} s + \Fg(X^{\thg}(T))$, and read off $\nabla_{\thg}J$ from the augmented adjoint.
\end{enumerate}
The second route makes the adjoint of \citetext{pmlr-v108-li20i} a tool for continuous-time RL: policy gradients can be computed using the same Virtual-Brownian-Tree machinery used for Neural SDE VAEs.

\subsection{The Exploratory Formulation of Wang et al. (2020)}

The deterministic-policy formulation above has a well-known shortcoming from an RL perspective: it offers no systematic mechanism for exploration. \citetext{JMLR:v21:19-144} proposed an \emph{exploratory} relaxation of the stochastic control problem in which the control is, at each time, a \emph{probability distribution} over actions rather than a single deterministic value, and the cost is augmented by an entropy term that rewards exploration.

Let $\pi_{t}(\cdot\mid x)$ be a (measurable) family of probability densities on $\mathcal{U}$, one for each $(t,x)$. The \emph{exploratory dynamics} are obtained by averaging the original drift and diffusion under $\pi$:
\beq\label{e:exploratory_sde}
\mathrm{d} X(t) \;=\; \tilde b\bigl(X(t), \pi_{t}\bigr)\,\mathrm{d} t
\;+\; \tilde\sg\bigl(X(t), \pi_{t}\bigr)\,\mathrm{d} W(t)
\eeq
where
\beq\label{e:averaged}
\tilde b(x,\pi) = \int_{\mathcal{U}} b(x,u)\,\pi(\mathrm{d} u\mid x),
\qquad
\tilde\sg\tilde\sg^{\top}(x,\pi) = \int_{\mathcal{U}}\sg\sg^{\top}(x,u)\,\pi(\mathrm{d} u\mid x)
\eeq
The matrix square root $\tilde\sg$ of the averaged diffusion matrix is chosen measurably. The entropy-regularised cost is
\beq\label{e:explor_cost}
J^{\la}(\pi) \;=\; \mathbb{E}\!\lb\int_{0}^{T}\!\lp
\int_{\mathcal{U}} L(X,u)\,\pi_{s}(\mathrm{d} u\mid X)
+ \la \int_{\mathcal{U}} \log\pi_{s}(u\mid X)\,\pi_{s}(\mathrm{d} u\mid X)
\rp\mathrm{d} s + \Fg(X(T))\rb
\eeq
where $\la>0$ is the exploration temperature. The classical Wang--Zhou result for linear--quadratic problems is that the optimal $\pi^{\star}_{t}$ is \emph{Gaussian} with mean determined by the classical LQ optimal control and variance determined by $\la$; more generally the optimal exploratory policy is a Gibbs distribution proportional to $\exp(-\bar H/\la)$, where $\bar H$ is an appropriately defined Hamiltonian evaluated against the value function of the exploratory problem.

\subsection{The Exploratory Backward Stratonovich SDE}
\label{s:ebsde}

The natural question is: \emph{what is the analog of the Li et al.\ stochastic adjoint for the exploratory controlled SDE~\refeq{exploratory_sde}?} Answering it gives what we call an {\bf Exploratory Backward Stratonovich SDE} (EB-SSDE).

\medskip\noindent{\it Setup.}\;
Fix an exploratory policy $\pi_{\thg}$ with parameters $\thg$ (e.g.\ a neural network outputting the parameters of a distribution on $\mathcal{U}$). Write the exploratory SDE
\refeq{exploratory_sde} in Stratonovich form:
\beq\label{e:explor_strat}
\mathrm{d} X(t) \;=\; \tilde b_{\thg}\bigl(X(t)\bigr)\,\mathrm{d} t
\;+\; \tilde\sg_{\thg}\bigl(X(t)\bigr)\circ \mathrm{d} W(t)
\eeq
where $\tilde b_{\thg}$ includes the It\^o--Stratonovich correction of $\tilde b$, and the drift and diffusion now depend on $\thg$ through the averaging operation. Define the loss functional
\beq\label{e:explor_loss}
\mathcal{L}(\thg) \;=\;
\int_{0}^{T}\!\lp\bar L\bigl(X(s), \pi_{\thg,s}\bigr)
+ \la\,\mathcal{H}\bigl(\pi_{\thg,s}\bigr)\rp\mathrm{d} s
+ \Fg\bigl(X(T)\bigr)
\eeq
where $\bar L(x,\pi) = \int L(x,u)\,\pi(\mathrm{d} u\mid x)$ and $\mathcal{H}(\pi) = \int \log\pi\,\mathrm{d}\pi$.

\medskip\noindent{\it The EB-SSDE.}\;
Then~\refeq{explor_strat} with loss~\refeq{explor_loss} will yield an adjoint state $a(t) = \partial\mathcal{L}/\partial X(t)$ satisfying the backward Stratonovich SDE
\beq\label{e:ebsde}
\boxed{\;
\mathrm{d} a(t) \;=\; -\,a(t)^{\top}\,\partial_{x}\tilde b_{\thg}\bigl(X(t)\bigr)\,\mathrm{d} t
\;-\; a(t)^{\top}\,\partial_{x}\tilde\sg_{\thg}\bigl(X(t)\bigr)\circ \mathrm{d} W(t)
\;-\; \partial_{x}\bar L\bigl(X(t),\pi_{\thg,t}\bigr)\,\mathrm{d} t
\;}
\eeq
with terminal condition $a(T) = \Fg_{x}(X(T))$, and solved backward along the \emph{same} Brownian path used for the forward solve. The parameter gradient then becomes
\beq\label{e:ebsde_grad}
\nabla_{\thg}\mathcal{L} \;=\;
-\int_{0}^{T}\!\lb
a(t)^{\top}\partial_{\thg}\tilde b_{\thg}\bigl(X(t)\bigr)\,\mathrm{d} t
+ a(t)^{\top}\partial_{\thg}\tilde\sg_{\thg}\bigl(X(t)\bigr)\circ \mathrm{d} W(t)
+ \partial_{\thg}\!\lp\bar L + \la\mathcal{H}\rp\!\bigl(\pi_{\thg,t}\bigr)\mathrm{d} t
\rb
\eeq

\medskip\noindent{ We observe three characteristics:}
\begin{enumerate}
\item {\bf EB-SSDE is a pathwise backward SDE, not a Pardoux--Peng BSDE.} Like the Li et al.\ adjoint, it reuses the forward Brownian path rather than introducing a new martingale-representation term. This is a computational advantage: we inherit the Virtual-Brownian-Tree trick and $\mathcal{O}(\log L)$ memory.
\item {\bf The entropy term appears as a direct $\thg$-gradient, not as a backward-SDE source.} Because the entropy $\mathcal{H}(\pi_{\thg})$ depends on $\thg$ only through the policy (not through the state), it contributes a standard pathwise gradient and does not enter the drift of the adjoint~\refeq{ebsde}. This decoupling makes our EB-SSDE a clean generalization of the Li et al.\ adjoint: the classical adjoint is recovered in the limit $\la\to 0$ with the policy pinched to a delta function at a deterministic action.
\item {\bf The connection to Wang--Zhou.} The optimality condition $\nabla_{\thg}\mathcal{L} = 0$, together with standard variational calculus, recovers the Gibbs form $\pi^{\star}(u\mid x)\propto \exp(-\bar H(x,u;a)/\la)$ for the optimal exploratory policy, where $\bar H$ is the Hamiltonian built from the learned adjoint $a$. This makes the EB-SSDE a constructive, gradient-based route to exploratory policies that does not pass through solving an HJB PDE.
\end{enumerate}

\section{Extensions and Future Directions}
\label{s:future}

Our immediate efforts will focus on three primary objectives: 
(i)~solidifying the theoretical foundation by deriving~\refeq{ebsde} for an exploratory linear--quadratic control problem, numerically verifying that Gibbs policies optimized via $\thg$ converge to the closed-form Gaussian solutions in \citetext{JMLR:v21:19-144}; (ii)~providing a high-performance implementation of EB-SSDE within the {\tt torchsde} ecosystem, utilizing the Virtual Brownian Tree for memory-efficient, reproducible path synthesis; and (iii)~formalizing the astrophysical inference problem \cite{Fagin_2024} as a reinforcement learning task, where the ``control'' represents physical perturbations to a latent SDE, optimized for light-curve reconstruction fidelity.

Beyond these immediate goals, several promising research trajectories remain:

\begin{enumerate}
\item {\bf Non-Gaussianity and Jump-Diffusion Processes.} To better capture the heavy-tailed variability characteristic of certain AGN, we will extend the framework to include L\'evy-driven SDEs or jump-diffusion models of \citetext{JiaBenson2019}. This necessitates a re-derivation of the adjoint sensitivity equations, as the introduction of jumps breaks the standard continuity assumptions of the previously defined adjoint SDE in \citetext{pmlr-v108-li20i}.
\item {\bf Theoretical Decision Boundaries for Delay-Coupling.} The magnitude of $\Lambda$ dictates a critical trade-off between convergence stability and exploration. A highly promising future direction is to build rigorous mathematical theories that determine the exact decision boundary of $\Lambda$ (whether characterized by its magnitude in norm or its specific distribution). For instance, robust decision boundary in thr form of $\|\Lambda\| < \delta_0$, and $\|\Lambda\| \ge \delta_1$ might be find with uncertainty intervals. Establishing these boundaries helps define the conditions under which the Continuous-DMSGD algorithm is guaranteed to behave convergently versus when it is prone to divergence.
\item {\bf Reinforcement Learning Benchmarking.} To move beyond astrophysical toy cases, we will subject EB-SSDE to standard continuous-control benchmarks. This will provide a rigorous evaluation of whether the pathwise backward Stratonovich adjoint can compete with established actor--critic methods in high-dimensional state spaces.
\item {\bf Bayesian Uncertainty Quantification.} We plan to treat the drift and diffusion networks ($\thg$ and $\phi$ in~\refeq{neural_sde}) as stochastic variables. By employing Langevin-style samplers, we can disentangle the aleatoric uncertainty inherent in the dynamical system from the epistemic uncertainty of the model itself, providing a more robust measure of confidence in physical parameter estimation.
\end{enumerate}

\section*{Acknowledgments}
The authors would like to thank the mathematician \textit{Dr. Farzad Sabzikar} for his invaluable guidance during weekly meetings from \textit{2026/3} to \textit{2026/5}. He helped finding relevant and useful literature and provided key insights into the use of gradient descent with a delayed term.

\newpage

\bibliographystyle{oxford3}
\bibliography{ref}

\end{document}